\documentclass[sigconf]{acmart}
\usepackage{multirow}

\usepackage{amssymb}
\usepackage{pifont, paralist}
\usepackage{caption}
\usepackage{subcaption,graphicx}
\newcommand{\cmark}{\ding{51}}
\newcommand{\xmark}{\ding{55}}
\AtBeginDocument{%
  }

\copyrightyear{2023}
\acmYear{2023}
\setcopyright{acmlicensed}\acmConference[CIKM '23]{Proceedings of the 32nd ACM International Conference on Information and Knowledge Management}{October 21--25, 2023}{Birmingham, United Kingdom}
\acmBooktitle{Proceedings of the 32nd ACM International Conference on Information and Knowledge Management (CIKM '23), October 21--25, 2023, Birmingham, United Kingdom}
\acmPrice{15.00}
\acmDOI{10.1145/3583780.3615470}
\acmISBN{979-8-4007-0124-5/23/10}

\begin{document}

\title{DoRA: Domain-Based Self-Supervised Learning Framework for Low-Resource Real Estate Appraisal}


\author{Wei-Wei Du}
\email{wwdu.cs10@nycu.edu.tw}
\orcid{0000-0002-0627-0314}
\affiliation{
  \institution{Department of Computer Science, National Yang Ming Chiao Tung University}
  \streetaddress{}
  \city{Hsinchu}
  \state{}
  \country{Taiwan}
  \postcode{300}
}

\author{Wei-Yao Wang}
\email{sf1638.cs05@nctu.edu.tw}
\orcid{0000-0002-6551-1720}
\affiliation{
  \institution{Department of Computer Science, National Yang Ming Chiao Tung University}
  \streetaddress{}
  \city{Hsinchu}
  \state{}
  \country{Taiwan}
}

\author{Wen-Chih Peng}
\email{wcpeng@cs.nycu.edu.tw}
\orcid{0000-0002-0172-7311}
\affiliation{
  \institution{Department of Computer Science, National Yang Ming Chiao Tung University}
  \streetaddress{}
  \city{Hsinchu}
  \state{}
  \country{Taiwan}
}
\begin{abstract}
The marketplace system connecting demands and supplies has been explored to develop unbiased decision-making in valuing properties.
Real estate appraisal serves as one of the high-cost property valuation tasks for financial institutions since it requires domain experts to appraise the estimation based on the corresponding knowledge and the judgment of the market.
Existing automated valuation models reducing the subjectivity of domain experts require a large number of transactions for effective evaluation, which is predominantly limited to not only the labeling efforts of transactions but also the generalizability of new developing and rural areas.
To learn representations from unlabeled real estate sets, existing self-supervised learning (SSL) for tabular data neglects various important features, and fails to incorporate domain knowledge.
In this paper, we propose DoRA, a \textbf{Do}main-based self-supervised learning framework for low-resource \textbf{R}eal estate \textbf{A}ppraisal.
DoRA is pre-trained with an intra-sample geographic prediction as the pretext task based on the metadata of the real estate for equipping the real estate representations with prior domain knowledge.
Furthermore, inter-sample contrastive learning is employed to generalize the representations to be robust for limited transactions of downstream tasks.
Our benchmark results on three property types of real-world transactions show that DoRA significantly outperforms the SSL baselines for tabular data, the graph-based methods, and the supervised approaches in the few-shot scenarios by at least 7.6\% for MAPE, 11.59\% for MAE, and 3.34\% for HR10\%.
We expect DoRA to be useful to other financial practitioners with similar marketplace applications who need general models for properties that are newly built and have limited records.
The source code is available at https://github.com/wwweiwei/DoRA.
\end{abstract}

\ccsdesc[500]{Computing methodologies~Machine learning}
\ccsdesc[500]{Computing methodologies~Artificial intelligence}
\ccsdesc[500]{Applied computing~Electronic commerce}

\keywords{few-shot learning, self-supervised learning, real estate appraisal}


\maketitle

\section{Introduction}
\begin{figure}[h]
  \vspace{-3pt}
  \centering
 \includegraphics[width=0.7\linewidth]{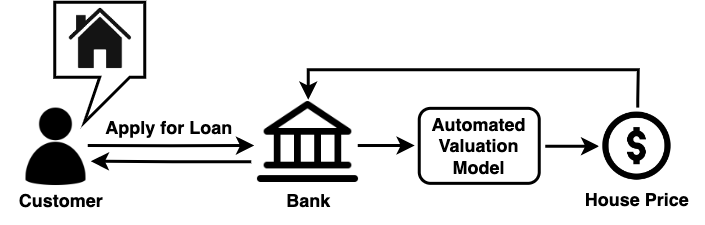}
  \vspace{-8pt}
  \caption{Illustration of the mortgage loan application process and how the AVM model works.}
  \label{fig:intro}
\end{figure}

\vspace{-5pt}
The exploration of property valuations has broad applicability across various domains.
Whether it involves developing strategies for mortgage lending, house rental, or security price revaluation, these scenarios can be effectively framed as property valuation systems characterized by intricate and high-cost domain knowledge.
In real estate appraisal, appraisers spend several hours estimating an individual property based on their knowledge \cite{human_valuation}, which introduces subjective biases of human estimations due to different understandings of the market \cite{DBLP:journals/eswa/AhnBOK12}.
Recently, automated valuation models (AVMs) including machine learning \cite{lin2011predicting,DBLP:conf/gecco/AzimluRM21} and graph-based approaches \cite{DBLP:conf/kdd/0003LZZLD021,DBLP:journals/corr/abs-2212-12190} have been developed to solve this issue by conducting a price estimation according to the information of real estate, as shown in Figure \ref{fig:intro}.

However, existing work adopting labeled datasets for supervised learning requires a large number of annotated labels and suffers from the generalization from seen to newly built appraisals.
In real-world scenarios, another challenging constraint is that properties are often sparse or newly built in most areas.
Beyond annotation costs, endlessly training specialized models on new types of real estate is not scalable in many practical scenarios.
Therefore, it is desirable to have a systematic approach to learn generic knowledge from existing unlabeled types of transactions to achieve effective quality with only very few annotated examples.
This problem is often defined as few-shot learning.
We note that defining \textit{low-resource scenarios} can be dependent on the goals and expectations of financial institutions and customers.
For instance, it can be defined as the number of transactions per city, per property type, or the combination of different factors. 
In this paper, we aim to tackle the actual application scenarios of few-shot real estate appraisal: \textbf{rural area}, \textbf{new developing area}, and \textbf{reducing label effort}.
Therefore, we define low-resource scenarios as the number of transactions per city.
However, how to effectively utilize unlabeled records with a high variability of property types remains a challenging problem.

The rapid development of self-supervised learning (SSL) has demonstrated the remarkable power of learning representation by using temporal information in spatial structure in images \cite{DBLP:conf/cvpr/He0WXG20,DBLP:conf/icml/ChenK0H20}, and semantic relationships in languages \cite{DBLP:journals/corr/abs-1301-3781,DBLP:conf/naacl/DevlinCLT19}.
This is beneficial in few-shot scenarios from the generic knowledge of pre-trained models.
However, the datasets of real estate appraisal are tabular domains which cannot directly apply these SSL techniques due to the different natures of their compositions (e.g., the 2D structure between pixels and the semantics between words).
Thus, several SSL approaches have been proposed to learn the general latent representation of tabular data \cite{DBLP:conf/aaai/ArikP21,DBLP:conf/nips/UcarHE21}.
Nonetheless, there is no existing approach that is able to integrate expert knowledge into the SSL objective for property estimation.
Existing tabular-based SSL methods are feature-agnostic, which does not consider the meaning of the features and regards all features as having the same importance.

In this paper, we propose DoRA, a domain-based self-supervised learning framework to tackle the low-resource scenarios for real estate appraisal.
In order to design an upstream task that produces universal representations, a pre-trained stage is introduced by solving the intra-sample domain-based pretext task (i.e., learning the appraiser's knowledge) from the unlabeled set. 
In addition, inter-sample contrastive learning is proposed to distinguish the similarities and discrepancies between transactions across towns.
In these manners, pre-trained embeddings generate robust representations in a lower dimension that contains more structured and domain-based information to use in the downstream task with only limited data.
In the fine-tuning stage, the pre-trained embedders and encoder are reused as a feature extractor for converting downstream transactions with pre-trained embeddings, and the weights are then adjusted based on a few target examples.

To summarize, the contributions of our work are as follows:
\begin{compactitem}
    \item To the best of our knowledge, DoRA is the first work focusing on low-resource real estate appraisal, which not only meets the needs of real-world scenarios but can also be adopted in other property valuations (e.g., house rental).
    \item The proposed framework is introduced with novel and effective intra- and inter-sample SSL objectives to learn robust geographical knowledge from unlabeled records.
    \item Extensive experiments were conducted to empirically show that DoRA is effective in few-shot settings compared with existing methods. We also illustrate a developed system of DoRA and the real-world industrial scenarios for cities and towns with extremely limited transactions.
\end{compactitem}
\vspace{-9pt}
\section{Related Work}
\noindent\textbf{Real Estate Appraisal.}
Previous works defined real estate appraisal as a supervised regression problem, and addressed it with machine learning techniques \cite{DBLP:conf/icdm/GeWXLZ19,DBLP:conf/kdd/0003LZZLD021,DBLP:journals/corr/abs-2212-12190}.
To incorporate multi-modal data sources for improving the performance, \citet{DBLP:conf/ssci/ZhaoCT19} took the visual content of rooms into account using a deep learning framework with XGBoost \cite{DBLP:conf/kdd/ChenG16}, while \citet{DBLP:journals/mta/BinGLL19} utilized street map images with attention-based neural networks.
On the other hand, LUCE was proposed to tackle spatial and temporal sparsity with the lifelong learning heterogeneous information network consisting of graph convolutional networks and long short-term memory networks \cite{DBLP:journals/corr/abs-2008-05880}.
Nonetheless, utilizing unlabeled transactions with high variability for low-resource real estate appraisal remains an unexplored yet challenging problem, which is also beneficial for property valuations.
We, therefore, aimed to design a self-supervised learning approach to learn domain representations of transactions from the unlabeled set.

\noindent\textbf{SSL in Tabular Data.}
Recently, SSL has achieved prominent success in the image \cite{DBLP:conf/cvpr/He0WXG20,DBLP:conf/icml/ChenK0H20}, audio \cite{DBLP:journals/spl/TagliasacchiGQR20}, and text \cite{DBLP:journals/corr/abs-1301-3781,DBLP:conf/naacl/DevlinCLT19} research fields.
However, these approaches are often difficult to transfer to the tabular domain since tabular data do not have explicit structures to learn the contextualized representations.
Therefore, multiple SSL approaches are proposed to learn the relation and latent structure between features in the tabular data domain \cite{DBLP:conf/acl/YinNYR20,DBLP:conf/nips/YoonZJS20,DBLP:conf/aaai/ArikP21,DBLP:conf/nips/UcarHE21}.
However, there is another domain that has not yet been explored: SSL for property valuations, which is mainly comprised of records in tabular format, and requires domain knowledge to evaluate the property objectively.
To that end, we have designed a pretext task based on domain knowledge and inter-sample contrastive learning to reinforce the model equipping domain-based contextualized representations of limited transactions for downstream tasks.
\vspace{-6pt}
\section{Preliminaries}



\subsection{Datasets}
\label{datasets}
Our three datasets (building, apartment, and house) were collected from the Taiwan Real Estate Transaction Platform \cite{datasource}, which contains real estate transactions in Taiwan from 2015 to 2021.
It is noted that the dataset of previous work \cite{DBLP:journals/corr/abs-2212-12190} is a subset of our collected building dataset since the authors adopted only the top 6 largest cities, while we expanded the dataset to all 22 cities and 3 different property types to accommodate more challenging but valuable real-world scenarios, and investigated the robustness of the model, especially for handling limited transactions.
Following \cite{DBLP:journals/corr/abs-2212-12190}, 3 months of transactions were used as the testing set (Apr. 2021 to Jun. 2021, 24,142 records), 3 months of transactions were used as the validation set (Jan. 2021 to Mar. 2021, 40,124 records), and the others were used as the training set.
We note that most real estate in the training set does not have corresponding prices due to the maintenance of the transaction platform; therefore, we used these unlabeled cases to form the unlabeled set (434,243 records).
We also collected two additional types of features to comprehensively describe the real estate from a neighboring and global view: PoI (Point of interest), and economic and geographical features.
We summarize these features as follows due to space limitations.

\noindent\textbf{Real Estate Features.}
Real estate has 39 features, including 16 categorical features and 23 numerical features to represent the metadata, for instance, location, the layout of the real estate, the current condition of the real estate, and household facilities.



\noindent\textbf{PoI Features.}
\label{poi_converter}
The original data is in spatial distribution format.
It was collected from a third-party map information company.
We designed a PoI converter to transform the data into a tabular format by dividing the facilities into YIMBY (Yes In My Back Yard, i.e., desirable adjacent public facilities) facilities, e.g., park, school, and NIMBY (Not In My Back Yard, i.e., non-desirable adjacent public facilities) facilities, e.g., power station, landfill. 
Then, the number of PoI of the real estate property was calculated with the Euclidean distance.
For instance, the feature YIMBY\_100 denotes the number of YIMBY facilities within 100 meters.


\noindent\textbf{Economic and Geographical Features.}
We added 7 external socio-economic features based on the real estate transaction quarter to represent the global view, including the house price index, unemployment rate, economic growth rate, lending rate, land transactions count, average land price index, and steel price index.
We also incorporated the land area and population density by the town name to consider the number of residents and the demand.


\subsection{Problem Formulation}


As discussed in the Introduction, we randomly sampled 1 and 5 shots for each city from the labeled training set as annotated examples (i.e., support set) to estimate the value of the real estate, and each example consisted of real estate features, PoI features, and economic and geographical features.
For instance, there is a total of 110 transactions in the 5-shot setting since the number of cities in Taiwan is 22.
We note that one of the cities, Lianjiang, does not have any apartments due to the nature of the city; thus, the apartment dataset only has 105 instances in the 5-shot setting.
It is noted that reporting each setting in the paper may not be feasible due to the page limit; therefore, we follow the standard few-shot settings (e.g., \cite{DBLP:conf/acl/MaBDAMAR22}) to report the overall performance in the experiments.
\section{Method}
\begin{figure}
  \centering
  \includegraphics[width=0.93\linewidth]{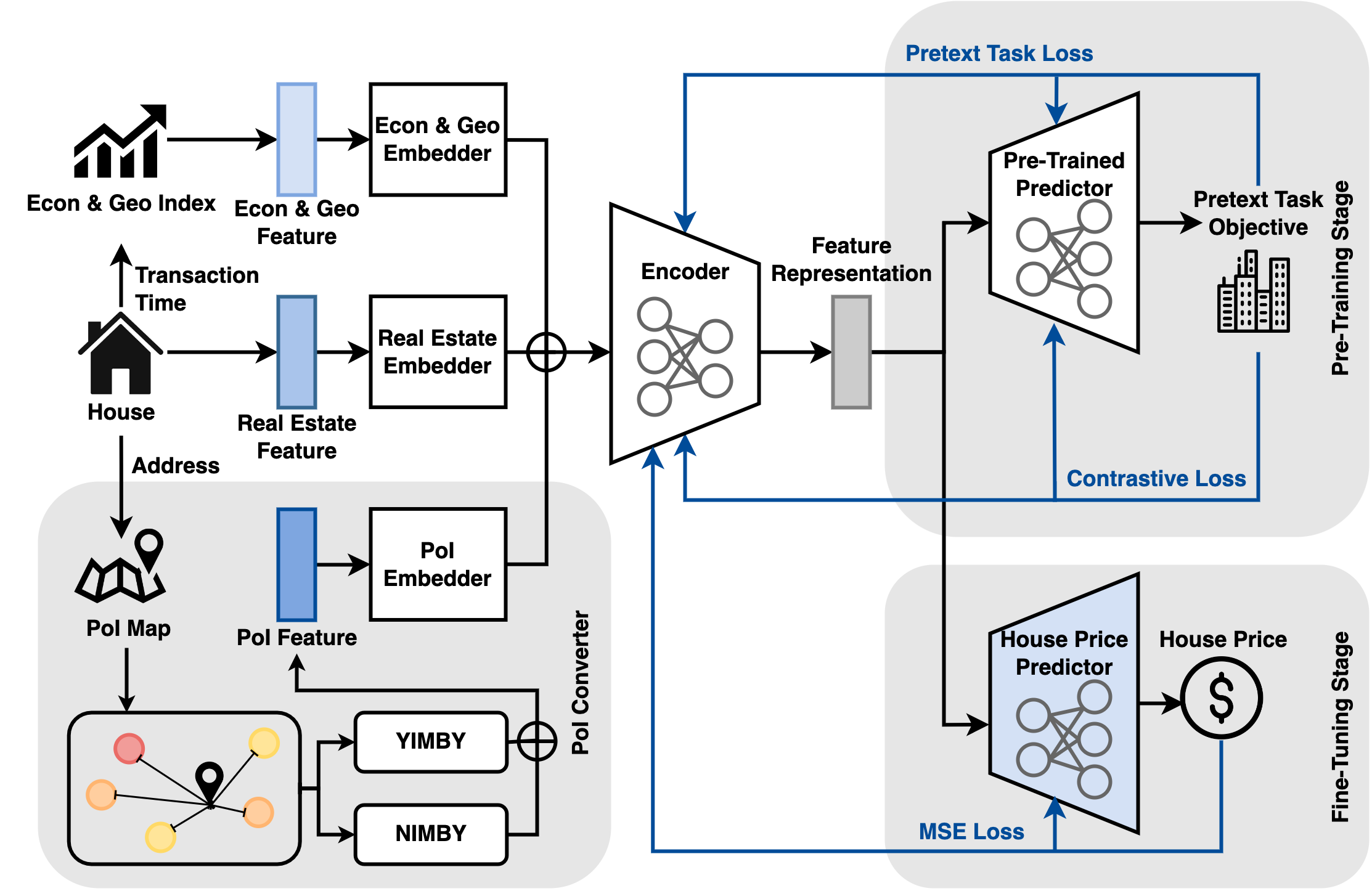}
  \caption{The framework of DoRA. The left side shows the three input data sources, including economic and geographical features, real estate features, and PoI features. The right side is the two stages of DoRA: 1) The pre-training stage for located town prediction and 2) The fine-tuning stage for real estate appraisal.}
  \vspace{-10pt}
  \label{fig:pretraining}
\end{figure}

The DoRA framework is illustrated in Figure \ref{fig:pretraining}.
DoRA takes economic and geographical features, real estate features, and PoI features as inputs.
The architecture of DoRA is mainly comprised of the embedder, encoder, and predictors.
The training pipeline is decomposed into two phases: \textbf{Pre-training stage:} Train by intra-sample pretext task and inter-sample contrastive learning to learn contextualized representations. \textbf{Fine-tuning stage:} Train with labeled data to predict the values of real estate.

\subsection{Model Architecture}
\noindent\textbf{Embedder.}
The heterogeneous input features can be categorized into 4 types: numerical real estate features, categorical real estate features, numerical economic and geographical features, and numerical PoI features.
The numerical real estate features $NR_i$ of $i$-th real estate are embedded as follows:
\begin{equation}
\small
    E_{NR_i} = Embedder_{NR}(NR_i),
\end{equation}
where the dimensions of $E_{NR_i}$ and $NR_i$ are $d_{NR}$ and 23, respectively.

To encode categorical real estate features, a naive method is to represent with one-hot encodings, which become sparse for high dimensions of categories and fail to preserve contextual information across features.
Thus, each category of real estate features is encoded separately and then concatenated as the categorical real estate embeddings:
\begin{equation}
\small
    E^{j}_{CR_i} = Embedder^{j}_{CR}(CR_i^{j}),
\end{equation}
\begin{equation}
\small
    E_{CR_i} = E^1_{CR_i} \oplus E^2_{CR_i} \oplus ... \oplus E^{16}_{CR_i},
\end{equation}
where $j$ indicates the index of the 16 categorical real estate features, $\oplus$ is the concatenation operator, and the dimensions of $E^{j}_{CR_i}$ and $E_{CR_i}$ are $d_{CR}$ and $16d_{CR}$, respectively.

Similarly, two embedders with the same architecture are employed as the numerical real estate embedder to encode numerical economic and geographical features $Econ\&Geo_i$ and PoI features $PoI_i$ of the $i$-th real estate property, respectively:
\begin{equation}
\small
    E_{Econ\&Geo_i} = Embedder_{Econ\&Geo}(Econ\&Geo_i),
\end{equation}
\begin{equation}
\small
    E_{PoI_i} = Embedder_{PoI}(PoI_i),
\end{equation}
where the input dimensions of $Econ\&Geo_i$ and $PoI_i$ are 9 and 16, respectively, and the dimensions of $E_{Econ\&Geo_i}$ and $E_{PoI_i}$ are $d_{Econ\&Geo}$ and $d_{PoI}$.

The aforementioned embedders are composed of an MLP and a Mish activation function \cite{misra2019mish} similar to \cite{DBLP:journals/corr/abs-2212-12190}.
Afterwards, the embedding of the $i$-th real estate $E_i$ is then concatenated as:
\begin{equation}
\small
    E_i = E_{NR_i} \oplus E_{CR_i} \oplus E_{Econ\&Geo_i} \oplus E_{PoI_i},
\end{equation}
where $E_i \in R^{d_{NR}+16d_{CR}+d_{Econ\&Geo}+d_{PoI}}$.

\noindent\textbf{Encoder.}
To encode the $i$-th embedding of real estate, the encoder is introduced to learn the contextualized feature representation $Z_i$ during both the pre-training and fine-tuning stages:
\begin{equation}
\small
     Z_i = Encoder(E_i),
\end{equation}
where $Z_i$ is a $d_{Z}$ dimension vector.
Since existing work on SSL for tabular data mainly focuses on the strategies of pretext tasks \cite{DBLP:conf/nips/YoonZJS20,DBLP:conf/nips/UcarHE21}, we also adopt MLPs with $N$ layers to align the comparison, where the number of layers $N$ of the encoder is optimized to ensure it is able to perform competitive performance of the pretext task (Section \ref{rq1}).
The output of the encoder can then be used to learn prior knowledge from the pre-training stage and downstream tasks from the fine-tuning stage.

\subsection{Pre-Training DoRA}
\label{pre_training_DoRA}

\noindent\textbf{Intra-Sample Pretext Task: Located Town Prediction.}
To enrich the feature representations from the unlabeled set, an intuitive method for designing the pretext task is to add noise and reconstruct the input, e.g., \cite{DBLP:conf/nips/YoonZJS20}, which treats all features with the same importance but neglects the domain knowledge of the meaning of the features.
Inspired by a recent study \cite{DBLP:conf/nips/LeeLSZ21} on the conceptual connections of features between pretext and downstream tasks benefiting the downstream tasks, we introduce a domain-based pretext task: \textbf{predict the located town of the given real estate}, which can also be adopted in various geographic-related tasks (e.g., house rental).
The input features of the real estate do not include the located town to avoid cheating.
In this way, the model is equipped with fine-grained domain knowledge to distinguish what might be the composition of real estate for each city, which benefits downstream tasks with limited transactions.

\noindent\textbf{Pre-Trained Predictor.}
The pre-trained predictor takes the feature representation of the $i$-th real estate property as the input, and predicts the corresponding town $\hat{Y}_i \in \mathbb{R}^{N_Y}$ with an MLP and the softmax activation function, where the number of $N_Y$ is 350:
\begin{equation}
\small
     \hat{Y}_i = Predictor_{pretext}(Z_i).
\end{equation}

\noindent\textbf{Pre-Training Loss.}
During the pre-training stage, the embedders, encoder, and pre-trained predictor are jointly trained in the following optimization problem:
\begin{equation}
\small
    \min_{M} \mathbb{E} [\alpha \mathbb{L}_{ce}(y, \hat{y}), (1-\alpha) \mathbb{L}_{cl}(y, \hat{y})],
\end{equation}
\sloppy
where $\alpha$ adjusts the weight between the two losses, $y$ is the ground-truth of the located town, and $M = \{Embedder_{NR}, Embedder_{CR}, Embedder_{Econ\&Geo}, Embedder_{PoI}$, $Encoder, Predictor_{pretext}\}$.
$\mathbb{L}_{ce}$ is the cross-entropy loss:
\begin{equation}
\small
    \mathbb{L}_{ce} = -\sum_{i=1}^{|C|} Y_i log(\hat{Y}_i),
\end{equation}
where C denotes the number of the unlabeled set.

\noindent\textbf{Inter-Sample Contrastive Learning.}
Since cross-entropy is sensitive to noisy labels and lessens generalization performance \cite{DBLP:conf/nips/ZhangS18, DBLP:conf/nips/ElsayedKMRB18}, we extend contrastive learning (CL) to incorporate label information to consider the similarities and discrepancies between real estate across towns.
$\mathbb{L}_{cl}$ is defined as:
\begin{equation}
\small
    \mathbb{L}_{cl} = \sum_{i \in I}\frac{-1}{|P(i)|} \sum_{p \in P(i)} log\frac{exp(Z_{i} \cdot Z_{p}/\tau)}{\sum_{a\in A(i)}{exp(Z_{i} \cdot Z_{a}/\tau)}},
\end{equation}
where $P(i)$ is the set of indices of all positive pairs, $A(i)$ is the set of all positive and negative pairs, $\tau$ is a temperature parameter, and $I$ is the instances (i.e., real estate) in a batch.
Positive pairs represent the two instances located in the same town, while negative pairs represent the two instances located in different towns.
As a result, $Z_p$ is one of the contextualized embeddings that is sampled from the same town within the batch as the positive pairs for CL.
By optimizing $\mathbb{L}_{cl}$, embeddings with the same class are closer, and embeddings from different classes are pulled over.

\subsection{Fine-Tuning DoRA}
\noindent\textbf{House Price Predictor.}
To train a regressor for the house price prediction, the feature representations of the pre-trained encoder of the $i$-th real estate property are fed into the house price predictor:
\begin{equation}
\small
     \hat{P}_i = Predictor_{price}(Z_i),
\end{equation}
where $\hat{P}_i$ is the estimated price.

\noindent\textbf{Fine-Tuning Loss.}
To optimize the estimated prices of real estate, the pre-trained embedders, pre-trained encoder, and the house price predictor of DoRA are jointly fine-tuned by minimizing the mean square error loss $\mathbb{L}_{mse}$:
\begin{equation}
\small
    \mathbb{L}_{mse} = \frac{1}{|S|} \sum \limits_{s_{i}\in S}(\hat{P}_{i}-P_{i})^2,
\end{equation}
where $S$ is the support set and $P_{i}$ is the $i$-th ground truth price.
\vspace{-5pt}
\section{Experiments}
In this section, we attempt to answer the following research questions on three property types of real-world real estate datasets:
\begin{compactitem}
    \item [\textbf{RQ1}] How does DoRA perform in the few-shot real estate appraisal? (Section \ref{rq1})
    \item [\textbf{RQ2}] Do each proposed component, the pretext task, and the input features contribute to DoRA? (Section \ref{ablation})
    \item [\textbf{RQ3}] How can DoRA be deployed for low-resource real estate scenarios? (Section \ref{deployed_system})
\end{compactitem}

\begin{table*}
    \centering
    \small
    \caption{Overall 1-shot performance evaluated by MAPE, MAE, and HR10\% on the building, apartment, and house datasets.}
    \vspace{-8pt}
    \label{tab:1shot}
    \scalebox{0.85}{
    \newcommand{\specialcell}[2][c]{%
  \begin{tabular}[#1]{@{}c@{}}#2\end{tabular}}

\begin{tabular}{cc|ccc|ccc|ccc}
    \toprule
    & & \multicolumn{3}{c|}{Building} & \multicolumn{3}{c|}{Apartment} & \multicolumn{3}{c}{House}\\
    \cmidrule{3-11}
    Type & Model & MAPE ($\downarrow$) & MAE ($\downarrow$) & HR10\% ($\uparrow$) & MAPE ($\downarrow$) & MAE ($\downarrow$) & HR10\% ($\uparrow$) & MAPE ($\downarrow$) & MAE ($\downarrow$) & HR10\% ($\uparrow$) \\
    \midrule
    STA & HA  & 45.36$\pm$0.97 & 12.93$\pm$0.25 & \underline{13.91$\pm$0.09} &
                                              46.20$\pm$2.73 & 16.74$\pm$0.36 & 13.69$\pm$1.19 &
                                              43.40$\pm$0.84 & 7.74$\pm$0.04 & 18.29$\pm$0.19 \\
    \midrule
    \multirow{4}{*}{SUP}       & LR         & 213.1$\pm$151.6 & 33.95$\pm$24.52 & 4.61$\pm$4.42 &
                                              153.3$\pm$19.65 & 34.21$\pm$6.57  & 3.69$\pm$1.68 &
                                              167.6$\pm$40.80 & 27.23$\pm$6.66 & 4.24$\pm$2.94 \\
                & XGBoost                   & 50.46$\pm$11.12 & 13.22$\pm$3.24 & 11.42$\pm$3.75 &
                                              40.21$\pm$2.47  & \underline{14.17$\pm$2.22} & 16.86$\pm$1.70 &
                                              52.69$\pm$3.32  &  9.77$\pm$1.51 & 13.80$\pm$3.58 \\
                & \specialcell{DNN}         & 41.31$\pm$2.14  & 13.06$\pm$0.74 & 12.50$\pm$1.94 &
                                              40.84$\pm$2.96  & 15.72$\pm$1.27 & 13.56$\pm$2.52 &
                                              37.93$\pm$4.65  &  8.06$\pm$0.47 & 16.20$\pm$1.17 \\
                & \specialcell{DNN + CL}    & 41.68$\pm$2.41  & \underline{12.85$\pm$0.77} & 12.32$\pm$1.49 &
                                              38.34$\pm$1.41 & 14.47$\pm$2.04 & 14.68$\pm$2.88 &
                                              \underline{37.00$\pm$0.82} & 7.86$\pm$0.87 & 16.20$\pm$4.04\\
    \midrule
    \multirow{2}{*}{SSL} & DAE              & 44.67$\pm$0.90 & 12.93$\pm$1.47 & 12.51$\pm$0.68 &
                                              45.17$\pm$0.90 & 14.25$\pm$2.30 & 14.16$\pm$0.44 &
                                              38.40$\pm$2.81 &  \underline{6.78$\pm$0.14} & \underline{21.77$\pm$0.84} \\
                & SubTab     & \underline{39.20$\pm$3.49} & 22.32$\pm$7.30 & 12.96$\pm$3.46 &
                                              \underline{37.99$\pm$0.79} & 14.34$\pm$0.45 & \underline{17.44$\pm$0.81} &
                                              41.96$\pm$1.31 & 14.23$\pm$0.86 & 13.77$\pm$2.95\\
    \midrule
    \multirow{2}{*}{Graph} & MugRep & 52.28$\pm$12.03 & 22.74$\pm$3.44 & 11.53$\pm$3.63 & 51.62$\pm$13.52 & 22.04$\pm$8.52 & 12.22$\pm$1.80 & 51.38$\pm$10.86 &  21.87$\pm$7.12 & 11.65$\pm$4.69 \\
                & ReGram     & 41.83$\pm$1.31 & 14.47$\pm$3.65 & 12.77$\pm$3.30 & 42.36$\pm$1.25 & 13.09$\pm$3.31 & 14.90$\pm$0.86 & 39.13$\pm$0.31 &  7.94$\pm$0.27 & 16.94$\pm$1.06 \\
    \midrule
                & \specialcell{DoRA (Ours)} & \textbf{38.77$\pm$2.85} & \textbf{11.16$\pm$0.85} & \textbf{14.53$\pm$1.24} &
                                              \textbf{33.73$\pm$1.04} & \textbf{10.51$\pm$2.14} & \textbf{19.55$\pm$2.25} &
                                              \textbf{33.59$\pm$1.59} & \textbf{6.53$\pm$0.25} & \textbf{22.41$\pm$1.72} \\
    \bottomrule
\end{tabular}
    }
\end{table*}

\begin{table*}
    \centering
    \small
    \caption{Overall 5-shot performance evaluated by MAPE, MAE, and HR10\% on the building, apartment, and house datasets.}
    \vspace{-8pt}
    \label{tab:5shot}
    \scalebox{0.85}{
    \newcommand{\specialcell}[2][c]{%
  \begin{tabular}[#1]{@{}c@{}}#2\end{tabular}}

\begin{tabular}{cc|ccc|ccc|ccc}
    \toprule
    & & \multicolumn{3}{c|}{Building} & \multicolumn{3}{c|}{Apartment} & \multicolumn{3}{c}{House} \\
    \cmidrule{3-11}
    Type & Model & MAPE ($\downarrow$) & MAE ($\downarrow$) & HR10\% ($\uparrow$) & MAPE ($\downarrow$) & MAE ($\downarrow$) & HR10\% ($\uparrow$) & MAPE ($\downarrow$) & MAE ($\downarrow$) & HR10\% ($\uparrow$) \\
    \midrule
    STA & HA  & 45.17$\pm$0.97 & 13.04$\pm$0.24 & 13.78$\pm$0.09 &
                                              46.20$\pm$2.73 & 17.03$\pm$0.19 & 13.69$\pm$1.19 &
                                              45.74$\pm$2.66 &  7.77$\pm$0.04 & 18.47$\pm$0.25\\
    \midrule
    \multirow{4}{*}{SUP}        & LR       & 99.74$\pm$34.19 & 17.38$\pm$3.06 & 10.06$\pm$2.46 &
                                             127.0$\pm$97.30 & 30.59$\pm$15.23 & 8.58$\pm$3.98 &
                                             245.2$\pm$277.1 & 39.70$\pm$37.55 & 6.75$\pm$6.11\\
                & XGBoost                  & \underline{34.75$\pm$2.46}  &  \underline{8.63$\pm$0.82} & 18.76$\pm$4.65 &
                                             39.53$\pm$11.50 & 11.65$\pm$2.40 & 20.54$\pm$4.53 &
                                              36.29$\pm$8.74 &  7.59$\pm$1.58 & \underline{23.00$\pm$2.64}\\
                &\specialcell{DNN} & 37.71$\pm$1.17 & 11.07$\pm$1.09 & 14.46$\pm$1.63 &
                                              38.66$\pm$3.14 & 14.62$\pm$1.52 & 15.41$\pm$0.85 &
                                              \underline{33.74$\pm$0.62} & 7.30$\pm$0.71 & 17.24$\pm$4.37\\
                &\specialcell{DNN + CL} & 36.77$\pm$2.10 & 10.94$\pm$1.21 & 15.23$\pm$2.91 &
                                              37.55$\pm$3.46 & 13.31$\pm$0.92 & 16.17$\pm$1.83 &
                                              34.54$\pm$1.04 & 7.63$\pm$0.63 & 16.36$\pm$3.91\\
    \midrule
    \multirow{2}{*}{SSL} & DAE & 43.93$\pm$1.47 & 11.73$\pm$1.51 & 13.73$\pm$1.38 &
                                              43.73$\pm$2.15 & 12.86$\pm$0.88 & 15.62$\pm$1.04 &
                                              39.69$\pm$0.64 & \underline{7.17$\pm$0.89} & 19.96$\pm$3.82\\
                & SubTab     & 35.29$\pm$0.38 & 14.82$\pm$9.00 & \textbf{22.65$\pm$0.19} &
                                              \underline{33.48$\pm$1.94} & \underline{10.31$\pm$0.35} & \textbf{22.64$\pm$2.06} &
                                              40.60$\pm$2.50 & 11.64$\pm$0.12 & 21.91$\pm$0.28 \\
    \midrule
    \multirow{2}{*}{Graph} & MugRep & 52.12$\pm$10.32 & 18.06$\pm$1.33 & 9.80$\pm$2.95 & 48.22$\pm$7.87 & 18.50$\pm$3.08 & 11.45$\pm$3.07 & 47.05$\pm$9.25 &  11.70$\pm$4.77 & 14.11$\pm$4.27 \\
                & ReGram     & 40.58$\pm$0.41 & 13.02$\pm$2.50 & 13.21$\pm$1.61 & 41.19$\pm$1.71 & 14.86$\pm$3.33 & 14.21$\pm$0.38 & 37.07$\pm$1.86 &  8.13$\pm$0.39 & 14.82$\pm$6.09 \\
    
    \midrule
                & DoRA (Ours)              & \textbf{31.85$\pm$0.50} & \textbf{7.79$\pm$0.23} & \underline{20.16$\pm$0.64} &
                                              \textbf{30.27$\pm$0.31} & \textbf{9.82$\pm$0.15} & \underline{20.54$\pm$0.52} &
                                              \textbf{31.66$\pm$0.27} & \textbf{6.28$\pm$0.05} & \textbf{23.13$\pm$0.10}\\
    \bottomrule
\end{tabular}
    }
\end{table*}

\subsection{Experimental Setup}
\noindent\textbf{Implementation Details.}
The dimensions of $d_{NR}$, $d_{Econ\&Geo}$, and $d_{PoI}$ are 16, the dimension of $d_{CR}$ is 10, and the dimension of $d_{Z}$ is 256.
The number of layers of the encoder $N$ is 6 and is designed with hidden dimensions $(2d_{Z}, 4d_{Z}, 8d_{Z}, 4d_{Z}, 2d_{Z}, d_{Z})$ in order.
We set weight $\alpha$ as 0.7 and temperature $\tau$ as 0.1.
For the numerical features, we impute them with standard normalization.
In the fine-tuning stage, both DoRA and other baselines use an MLP as the house price predictor.
We employ the AdamW optimizer \cite{DBLP:conf/iclr/LoshchilovH19} using the learning rate of 0.005, and the batch size is 512.
During the pre-training stage, we use all property types of unlabeled sets to learn a pre-trained model to enforce the generalizability and then fine-tune based on various property type datasets.
The training epochs of the pre-training and fine-tuning stages are 150 and 200, respectively.
All of the hyper-parameters in the experiment were tuned based on the validation set.
All experiments were repeated 5 times with different random seeds for sampling support sets to reduce the bias of few-shot sampling and to report average metrics with standard deviations for each evaluation metric.

\noindent\textbf{Evaluation Metrics.}
Previous work mainly focused on regression metrics for evaluating house price prediction \cite{DBLP:conf/kdd/0003LZZLD021, DBLP:journals/corr/abs-2212-12190}.
We extended these metrics with hit rate k\% \cite{goodman2003housing} to measure the accuracy of target properties within a tolerance error percentage k conforming to real-world financial requirements.
Therefore, we adopted mean absolute percentage error (MAPE), mean absolute error (MAE), and hit rate 10\% (HR10\%) to comprehensively evaluate the results.





\noindent\textbf{Baselines.}
The baselines can be categorized into four groups: 1) \textbf{Statistics model (STA):} Historical Average (HA), 2) \textbf{Supervised models (SUP):} Linear Regression (LR), XGBoost \cite{DBLP:conf/kdd/ChenG16}, DNN, and DNN with contrastive learning (DNN + CL), 3) \textbf{Self-supervised models (SSL)}: DAE \cite{DBLP:conf/icml/VincentLBM08} and SubTab \cite{DBLP:conf/nips/UcarHE21}, and 4) \textbf{Graph-based models (Graph)}: MugRep \cite{DBLP:conf/kdd/0003LZZLD021} and ReGram \cite{DBLP:journals/corr/abs-2212-12190}.
The SSL baselines are also pre-trained using the unlabeled set and are then fine-tuned to the house price prediction task.

\subsection{Overall Performance}
\label{rq1}

The pre-training performance of DoRA reaches about 0.85 and 0.96 in terms of macro-f1 and micro-f1 scores to ensure that DoRA is capable of detecting the geographic locations of real estate.
Table \ref{tab:1shot} and Table \ref{tab:5shot} summarize the performance comparisons of various methods with 1-shot and 5-shot scenarios.
The best result of each metric is highlighted in boldface and the second best is underlined.
Quantitatively, the improvement in DoRA is at least 7.6\% for MAPE, 11.59\% for MAE, and 3.34\% for HR10\% on average for the three property types.
We make the following observations: 

1) DoRA consistently outperforms STA, SUP, SSL, and Graph approaches for the building, apartment, and house datasets with limited transactions, while some SUP models (e.g., XGBoost) are even inferior to the performance compared to the statistics-based method in the 1-shot scenario.
Moreover, SUP approaches significantly hinder the performance of all datasets, which indicates that supervised methods require a large amount of labeled data to achieve competitive performance.
We can also observe that graph-based methods considering the neighbor's information to construct a graph perform worse since most rural real estate does not have neighbors.
2) 
DAE and SubTab perform worse than DoRA in terms of all metrics and different scenarios, which verifies that randomly adding noise to the inputs and regarding all features with the same importance hinder models' learning of domain-based representations from the unlabeled set.
This also points out the attribution of taking advantage of the intra-sample domain-based pretext task in DoRA capable of improving downstream performance.
3) We also notice that adding contrastive loss to the DNN slightly improves some metrics, which implies that contrastive learning enriches representations for the downstream task, even when only applying limited data.

\begin{table}
    \small
    \caption{Model ablation with the building dataset. CL denotes contrastive learning.}
    \vspace{-5pt}
    \label{tab:ablation}
    \scalebox{0.95}{
    \newcommand{\specialcell}[2][c]{%
  \begin{tabular}[#1]{@{}c@{}}#2\end{tabular}}

\begin{tabular}{ccccc|c}
    \toprule
    \specialcell{Pre-Trained\\Datasets} & \specialcell{Pretext\\Task} & CL ($\alpha$) & $d_Z$ & \specialcell{Fine-Tune\\Encoder} & MAPE ($\downarrow$) \\
    \midrule
    \midrule
    Building & Town & + (0.7) & 256 & \cmark & 37.09 \\ 
    \midrule
    All & Town & + (0.7)  & 512 & \cmark & 33.65\\ 
    \midrule
    All & Town & + (0.7) & 256 & \xmark & 58.65\\
    \midrule
    All & Town & - & 256 & \cmark & 37.83\\ 
    \midrule
    All & Town & + (0.5) & 256 & \cmark & 36.37\\ 
    \midrule
    \midrule
    All & Town & + (0.7) & 256 & \cmark &  31.85 \\ 
    \bottomrule
\end{tabular}
    }
\end{table}


\begin{table*}
    \small
    \caption{The case study of 1-shot performance with three areas.}
    \vspace{-5pt}
    \label{tab:case_study_esun}
    \scalebox{0.85}{
    \newcommand{\specialcell}[2][c]{
  \begin{tabular}[#1]{@{}c@{}}#2\end{tabular}}

\begin{tabular}{cc|ccc|ccc|ccc}
    \toprule
    & & \multicolumn{3}{c|}{Building} & \multicolumn{3}{c|}{Apartment} & \multicolumn{3}{c}{House}\\
    \cmidrule{3-11}
    Area & Model & MAPE ($\downarrow$) & MAE ($\downarrow$) & HR10\% ($\uparrow$) & MAPE ($\downarrow$) & MAE ($\downarrow$) & HR10\% ($\uparrow$) & MAPE ($\downarrow$) & MAE ($\downarrow$) & HR10\% ($\uparrow$) \\
    \midrule
    \multirow{2.4}{*}{Nantun District, Taichung} & XGBoost  & 39.91 & 7.93 & 0.00 & 38.74 & 12.45 & 4.72 & 33.63 & 10.68 & 20.00 \\
    \cmidrule{3-11}
     & DoRA & \textbf{15.82} & \textbf{2.77} & \textbf{40.00} & \textbf{21.53} & \textbf{10.15} & \textbf{16.04} & \textbf{22.53} & \textbf{9.95} & \textbf{35.33} \\
    \midrule
    \midrule
    \multirow{2.4}{*}{Wugu District, New Taipei} & XGBoost & 54.63 & 11.74 & 16.67 & 65.11 & 16.62 & 0.00 & 162.85 & 28.53 & 0.00 \\
    \cmidrule{3-11} & DoRA  & \textbf{34.77} & \textbf{6.20} & \textbf{19.67} & \textbf{61.58} & \textbf{10.61} & \textbf{4.46} & \textbf{2.55} & \textbf{0.44} & \textbf{100.00} \\
    \midrule
    \midrule
    \multirow{2.4}{*}{Nantou City, Nantou County} & XGBoost & 52.69 & 6.86 & 7.69 & 61.58 & 10.61 & 4.46 & 34.86 & 7.52 & 10.14 \\
    \cmidrule{3-11} & DoRA  & \textbf{39.65} & \textbf{5.14} & \textbf{15.38} & \textbf{32.57} & \textbf{6.49} & \textbf{11.76} & \textbf{31.73} & \textbf{5.36} & \textbf{24.64} \\
    \bottomrule
\end{tabular}

    }
    \vspace{-3pt}
\end{table*}

\vspace{-25pt}
\subsection{Ablation Study}
\label{ablation}
\noindent\textbf{Model Ablation.}
We study a comprehensive component ablation with the building dataset in the 5-shot scenario in terms of MAPE. 
As shown in Table \ref{tab:ablation}, only using the unlabeled set of the corresponding type (building type in row 1) significantly degenerates the downstream performance, which signifies that DoRA is able to leverage the unlabeled set from various types to improve model performance on the house price prediction.
Rows 2, 3, 4, and 5 present the sensitivity analysis of different hyper-parameters, which shows that removing contrastive loss and changing the weight of $\alpha$ reduce the performance more substantially.
In addition, freezing the pre-trained encoder and using the different dimensions of feature representations also negatively impact the house price performance.

\begin{figure}
     \small
     \centering
     \begin{subfigure}[b]{0.22\textwidth}
         \centering
         \includegraphics[width=\linewidth]{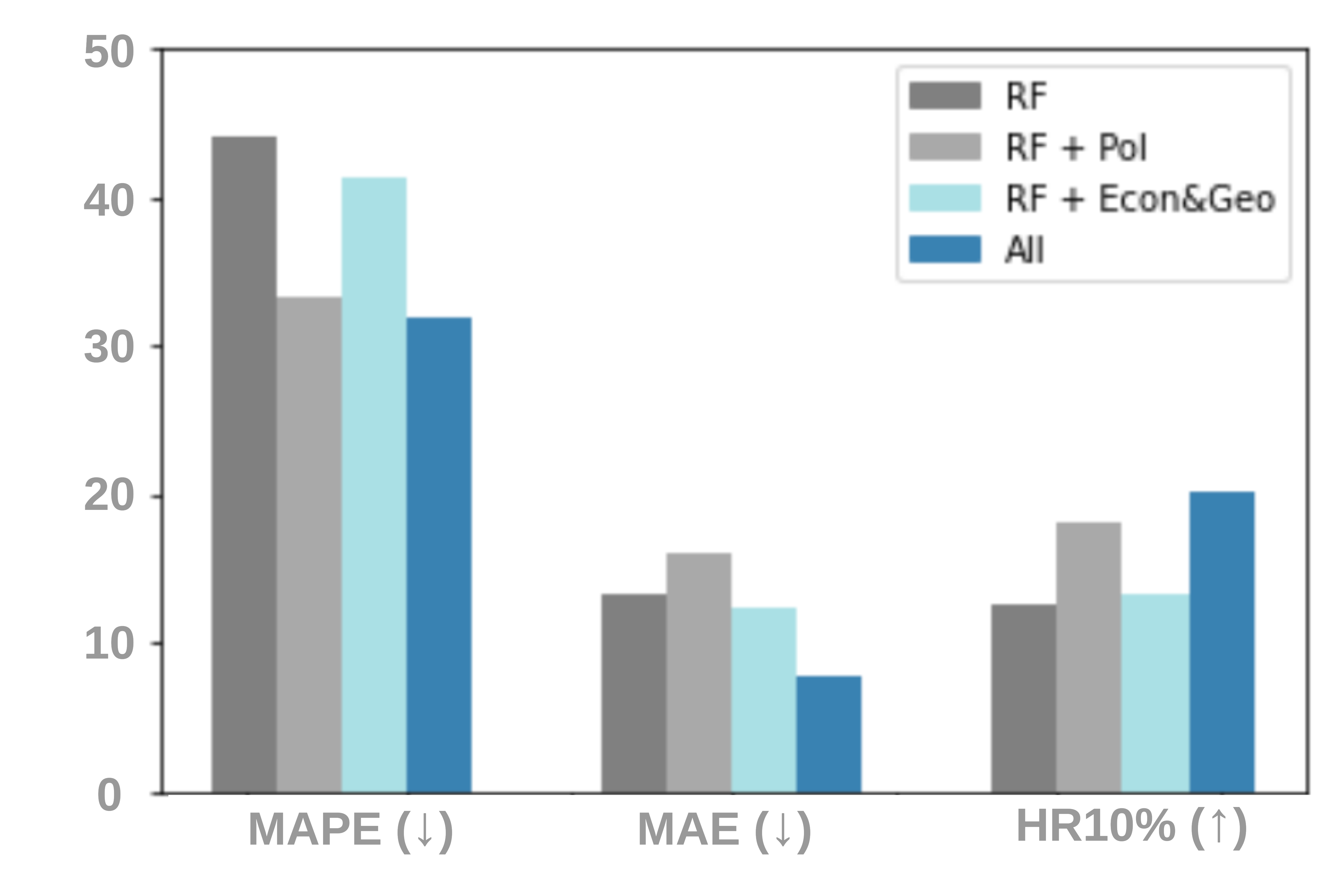}
         \vspace{-10pt}
         \caption{Different input features.}
         \label{fig:feature_ablation}
     \end{subfigure}
     \hfill
     \begin{subfigure}[b]{0.22\textwidth}
         \centering
         \includegraphics[width=\linewidth]{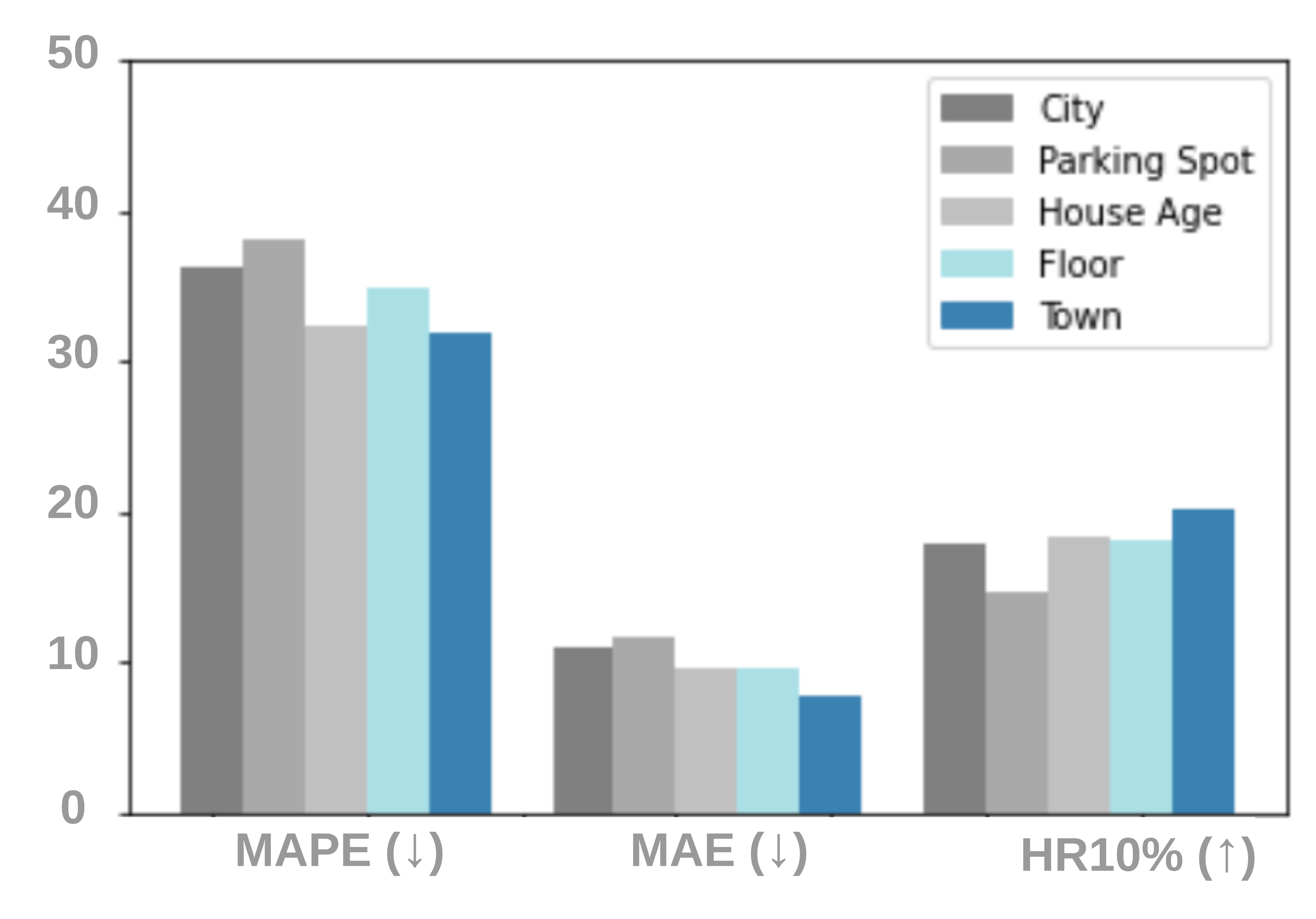}
         \vspace{-10pt}
         \caption{Different pretext tasks.}
         \label{fig:pretext_task}
     \end{subfigure}
        \vspace{-10pt}
        \caption{Feature and task ablations with the building dataset.}
        \label{fig:ablation}
    \vspace{-13pt}
\end{figure}

\noindent\textbf{Feature and Pretext Task Ablations.}
\label{feature_ablation}
To investigate the relative effects of different feature sources, we evaluate DoRA with full features and its three variants:
1) \textbf{RF} includes only real estate features, which is the basic feature to describe real estate;
2) \textbf{RF+PoI} includes both real estate features and PoI features;
3) \textbf{RF+Econ\&Geo} includes both real estate features and economic and geographical features;
4) \textbf{All} includes the complete set of features.
Figure \ref{fig:feature_ablation} reports the performance with the building dataset in the 5-shot setting.
Removing either PoI features or economic and geographical features leads to inferior performance in terms of all metrics.
Moreover, PoI features affect the performance more considerably compared with economic and geographical features, which indicates that the neighboring facilities are critical factors for real estate appraisal.
These observations suggest that considering only the metadata of real estate is insufficient for house price prediction, while various sources of describing real estate from a global viewpoint enhance the capability of the model.
We also examine various pretext tasks as shown in Figure \ref{fig:pretext_task}, which replaces the original pretext task with another pretext task from the real estate metadata.
We can observe that all metrics performance of replacing with different pretext tasks is degraded, which empirically showcases the importance of the pretext task objective.

\vspace{-4pt}
\subsection{Deployment}
\noindent\textbf{Deployed System.} 
\label{deployed_system}
E.SUN Bank is a commercial bank encouraging AI-driven solutions for businesses in Taiwan.
In the past, real estate appraisal tasks were primarily manual, high-cost, and subjective to the appraiser.
Moreover, it is challenging to estimate real estate if there are limited historical transactions.
To that end, we partnered with the fintech team to deploy automated DoRA to perform an online mortgage calculator platform, which requires real estate appraisal for suggesting the mortgage.
In the prototype system, the user is required to enter the information (e.g., address, house age, parking space, etc.) of the property that is to be mortgaged.
DoRA will then execute an online real estate appraisal by extracting the PoI features based on the house address as part of the inputs.
Afterwards, the appraised price will be incorporated with other internal entities to compute the approximate loan.

\noindent\textbf{Case Study.} 
\label{case_study}
As the distribution of transactions for each city is a long-tail distribution, the great majority of the cities only have a few transactions.
Therefore, we simulated three cities with extremely limited transactions and compared DoRA with XGBoost.
As shown in Table \ref{tab:case_study_esun}, we can observe that DoRA is significantly superior to the baseline for all metrics, particularly in property types where it fails to appraise real estate with 0 hit rate scores. 
The performance of the house type in Wugu District shows that DoRA appraises effectively while the baseline deteriorates all metrics more substantially.
These cases confirm that DoRA is capable of low-resource scenarios due to the incorporation of the unlabeled set.
In partnership with the fintech team, such an improvement definitely improves resource utilization and performance of the real estate appraisal.

\section{Conclusion}
In this work, we propose DoRA, a domain-based SSL framework for low-resource real estate appraisal, which is one of the challenging property valuation tasks due to heavy human efforts. 
Predicting the geographic location as an intra-sample pretext task reinforces the model learning the domain-based representations of real estate. 
Moreover, DoRA integrates inter-sample contrastive learning to distinguish the discrepancies between transactions across towns for the robustness of limited examples in downstream tasks.
Extensive results on different property types of real-world real estate appraisals demonstrate that DoRA consistently outperforms supervised, graph, and SSL approaches in few-shot scenarios.

Prior to this work, real estate estimations with new and rural transactions were mainly evaluated by appraisers using manual, ad-hoc, intuition-driven methods at E.SUN Bank.
Now, fintech teams have adopted DoRA for automatically estimating values of real estate, which saves time and drives objectivity in the business, and empowers the financial institution to plan and adapt dynamically to new information.
As our proposed approach was flexibly designed with SSL, we expect DoRA to be useful to other applied scientists in financial marketplaces, especially those whose goal is to perform property valuation with only a few labeled data.
\section{Acknowledgments}
We would like to thank Fu-Chang Sun, Yi-Hsun Lin, Hsien-Chin Chou, Chih-Chung Sung, and Leo Chyn from E.SUN Bank for sharing data and discussing the findings.

\bibliographystyle{ACM-Reference-Format}
\balance
\bibliography{reference}

\end{document}